\documentclass[5p,twocolumn]{elsarticle}
\usepackage{graphicx}
\usepackage{amsmath,amssymb} 
\usepackage{color}
\usepackage{amsmath,mathtools}
\usepackage[table]{xcolor}
\usepackage{multirow}
\usepackage{soul}
\usepackage{subcaption}
\usepackage{tabularx}
\usepackage[numbers]{natbib} 
\usepackage{epstopdf}
\usepackage{lineno}
\usepackage[hidelinks]{hyperref}
\usepackage{arydshln} 

\newcommand{\lst}[1]{\textcolor{black}{#1}}

\journal{.}

\biboptions{sort&compress}

\begin{document}  
	
	\begin{frontmatter}
		\title{Automatic Defect Segmentation on Leather with Deep Learning}
		
		\author[add1]{Sze-Teng Liong} 
		\ead{stliong@fcu.edu.tw}
		\author[add2]{Y.S. Gan\corref{cor1}}
		\ead{ysgn88@gmail.com}
		\author[add2]{Yen-Chang Huang} 
		\ead{yenchang.huang1@gmail.com}
		\author[add3]{Chang-Ann Yuan} 
		\ead{cayuan@fcu.edu.tw}
		\author[add1]{Hsiu-Chi Chang}
		\ead{aspirine923768@gmail.com}
		
		\cortext[cor1]{Corresponding author}
		\address[add1]{Department of Electronic Engineering, Feng Chia University, Taichung, Taiwan}
		\address[add2]{Research Center for Healthcare Industry Innovation, NTUNHS, Taipei, Taiwan} 
		\address[add3]{Department of Mechanical and Computer-Aided Engineering, Feng Chia University, Taichung, Taiwan}

		\begin{abstract}
			Leather is a natural and durable material created through a process of tanning of hides and skins of animals.
			The price of the leather is subjective as it is highly sensitive to its quality and surface defects condition.
			In the literature, there are very few works investigating on the defects detection for leather using automatic image processing techniques.
			The manual defect inspection process is essential in an leather production industry to control the quality of the finished products. 
			However, it is tedious, as it is labour intensive, time consuming, causes eye fatigue and often prone to human error.
			In this paper, a fully automatic defect detection and marking system on a calf leather is proposed.
			The proposed system consists of a piece of leather, LED light, high resolution camera and a robot arm.
			Succinctly, a machine vision method is presented to identify the position of the defects on the leather using a deep learning architecture.
			Then, a series of processes are conducted to predict the defect instances, including elicitation of the leather images with a robot arm, train and test the images using a deep learning architecture and determination of the boundary of the defects using mathematical derivation of the geometry.
			Note that, all the processes do not involve human intervention, except for the defect ground truths construction stage.
			The proposed algorithm is capable to exhibit \lst{91.5\%} segmentation accuracy on the train data and \lst{70.35\%} on the test data.
			We also report confusion matrix, F1-score, precision and specificity, sensitivity performance metrics to further verify the effectiveness of the proposed approach.

		\end{abstract}
		
		
	\end{frontmatter}
	
	\section{Introduction}
	
	Hides (refers to skin of large animals, i.e., cows) and skins (normally used for small animals, i.e., sheep) are mostly by-products of slaughterhouses.
	Many of the supply comes from the USA, Brazil and Europe, since they are large producer of beef.
	Normally, three major steps are carried out in a leather factory: sorting, chemical processing and physical processing.
	The raw materials are categorized by the number of defects, such as tick bites and scars which  affect in the quality degradation and subsequently require more processing.
	The materials are then tumbler with specific chemical substances that convert the hides or skins into leathers, and hence possess the superior characteristics of soft, pliable, water resistance and putrefaction.
	Finally the leathers are stretched and trimmed to portray its velvet appearance.
	Those leathers are then sold to the leather manufacturing companies to produce high-end leather goods, like bags, shoes and jackets.
	The companies usually carry out a few rounds of defects sorting and classification by the severity for different appearance regions of the visual appeal on the goods.
	
	One of the earliest research works on the leather is carried out by Yeh and Perng~\cite{yeh2001establishing}.
	They defined a reference standard to classify the leather into several grade levels based on the amount of usable region that are eligible to proceed for the manufacturing process.
	As the price of every piece of leather hinges on its grade, such guideline is established to minimize the disputes for each trade transaction. 
	In fact, sophisticated negotiation always incurs additional cost and argument between suppliers and purchasers.
	During the defect detection process in~\cite{yeh2001establishing}, a few experienced experts are involved to manually annotate and mark the defect region using a software package - Adobe Photoshop~4.0~\cite{photoshop}.
	This process that requires human effort is not reliable as it is highly dependent on individuals.
	Thus, a machine vision technique is necessary to reduce the cost (i.e., workload and time) of the defect annotation task.
	
	To date, most of the defect detection and annotation procedures in industries are still carried out by highly trained inspectors.
	In general, there are two visual inspection automation tasks: classification and instance segmentation.
	The former categorizes the type of the defect of the leather, such as cuts, tick bites, wrinkle, scabies and others; whereas the latter localizes the defect region and at the meantime annotate the type of the defect.
	Many previous works focus on the leather defect classification.
	In contrast, there are relative less researchers predict the precise position of the defects.
	It is also worth noting that in many of the experiments reported in the papers, a test image only contain one type of the defects or identify the presence of a single defect in a sample image~\cite{wong2018computer, tong2016differential, villar2011new}.

	Kwon et al.~\cite{kwon2004development} propose a framework to identify several defect types (i.e., hole, pin hole, scratch and wrinkle) based on the histogram of the pixel intensity values.
	They discover that the composition of the image pixels of a non-defective leather should portray standard normal distributions.
	For the hole defects, their Gaussian distribution for image pixels are usually concentrated at the brighter part (i.e., close to pixel value of 255).
	In contrast, the pin holes are having much more darker pixels (i.e., close to pixel value of 0). 
	Defects like scratch and wrinkle are normally present distinct patterns compared to the normal distribution.
	Then, the grade of the leather (i.e., A, B or C) would be determined based on the analysis result that refers to the density and the number of defects extracted. 
	
	On the other hand, an image processing technique - fuzzy logic, is employed in~\cite{krastev2004leather} to analyze the features set of the leather images to perform the surface defects recognition.
	The leather image is first loaded in grey level and represented using a histogram. 
	Specifically, a few statistical features such as the histogram range, histogram position, median, mean, variance, energy, entropy, contrast, etc. are calculated with maximal, minimal and average values.
	However, the sample size is small (i.e., images) and the procedure of the experiments is ambiguous.
	For instance, the explanation for the experiment configuration and the distribution of the training and testing sets to evaluate the proposed algorithm are not included in the paper.

	An automated machine vision system to detect a few defect types (i.e., open cut , closed cut and fly bite) is introduced by Villar et al.~\cite{villar2011new}.
	They utilize seven popular feature descriptors (i.e., first order statistics, contrast characteristics, Haralick descriptors, Fourier and Cosine transform, Hu moments with information about intensity, Local binary patterns and Gabor features) and a selection method to dynamically reduce the feature size.
	Then, a multilayer perceptron neural classifier is adopted to categorize the type of the defect.
	An overall of 94\% high classification accuracy on the test images is obtained.
	Note that the training and testing datasets composed a total of approximately 1800 sample images of 40 $\times$ 40 spatial resolution.  
	
	A similar defect categorization work is conducted by Pistori et al.~\cite{pistori2018defect}.
	Particularly, they tend to distinguish four types of defects: tick marks, cuts, scabies and brand marks made from hot iron, on both the raw hide and wet blue leathers.
	The former has more complex exterior that has various kinds of surface (i.e., textures, colors, shapes, thickness and even with serious defects), whereas the latter is a common type of leather that had been undergoing a tanning process which appears to be more noticeable to both the human and machine visual inspection.
	The features of the images are extracted using a popular texture analysis technique, namely Gray-scale Coocurrence Matrix (GLCM)~\cite{jobanputra2004texture,  singh2002spatial}.
	The proposed method is validated on a pre-built dataset, comprised of images from 258 pieces of raw hide and wet blue leather with 17 different defect types.
	As a result, a perfect classification result (i.e., 100\%) is achieved using Support Vector Machine (SVM)~\cite{suykens1999least} classifier.

	Another leather classification work is conducted by Pereira et al.~\cite{pereiraclassification}, which attempt to reduce the discrepancies between specialists' assessments on the goat leather quality in order to increase the productivity of quality classification using feature extractor and machine learning classifier.
	The proposed algorithm includes the processes of image acquisition, image preprocessing, features extraction and machine learning classification. 
	In brief, a new approach is introduced to extract features called Pixel Intensity Analyzer (PIA) which emerges as the most cost-effective method to the problem when used together with Extreme Learning Machines (ELM) classifier~\cite{huang2006extreme}. 
	However, the details for the experiment setup is absence, thus it may be difficult to duplicate the framework and compare to the other works.
	
	A recent work is carried out by Winiarti et al.~\cite{winiarti2018pre}, where they classify five types of leather by employing both the two types of feature extractors: handcrafted feature descriptors and deep learning architecture. 
	The leather types include the monitor lizard, crocodile, sheep, goat and cow.
	For the handcrafted representation, a fusion of statistical color features (i.e., mean, standard deviation, skewness and kurtosis) and statistical texture features (i.e., contrast, energy, correlation, homogeneity and entropy) are adopted.
	A pre-trained AlexNet is exploited as the deep learning structure. 
	The classification performance suggests that the deep learning method can better capture the characteristics of the leather, which exhibits an overall accuracy of 99.97\%.
	Note that, there is no defect classification involved in this paper as all the data are non-defective images.

	For the defect localization and segmentation tasks, one of the pioneer research works is conducted by Lovergine et al.~\cite{lovergine1997leather}.
	They detect and determine the defective areas using a black and white CCD camera. 
	Then, a morphological segmentation~\cite{giardina1988morphological, lee1996mathematical} process is applied on the collected images to extract the texture orientation features of the leather.
	A few of qualitative results are shown in the paper, however, the quantitative methods and numerical data to evaluate the proposed algorithm are absence.

	Later, Lanzetta and Tantussi~\cite{lanzetta2002design} suggest a laboratory prototype for trimming the external part of a leather. 
	The leather sample images are processed to determine the background and the defective areas by using binarization, opening and laplacian mask methods to find the trimming path.
	The proposed defect detection system successfully identify most of the defects on distinct leather types.
	However, the surface finish and color are still the main factors that could influence the outcome of inspection.
	Since there is no practical implementation of the proposed prototype, the qualitative and quantitative performances are not shown in the paper.

	A 6-step inspection method has been proposed by Bong et al.~\cite{bong2018vision} for leather defect (i.e., scars, scratches and pinholes) detection and predict their exact size, shape and position.
	An image grabbing system is built such that the leather is able to captured using a static camera and a consistent light source.
	Then a series of image processing techniques are applied on the images captured to obtain the defective areas, which include color space conversion, Gaussian thresholding, Laplacian detection, Median blurring, defect features extraction (i.e., color moments, color corellogram, zernike moments, texture features) and SVM classification.
	The distribution of the training and evaluation data for the images that contain defects are about 7.5~:~2.5. 
	The proposed method achieves an average accuracy of 98.8\% to detect a single defect in every image.

\begin{figure}[t!]
		\centering
		\includegraphics[width=1\linewidth]{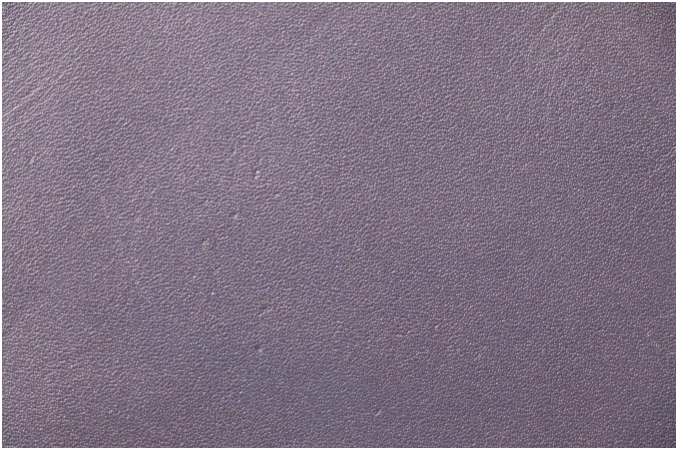}
		\caption{A sample defective image}
		\label{fig:sample_capture}
	\end{figure}

	The goal of this study is to introduce an automatic defect identification system to segment irregular regions of a specific defect type, viz., tick bites.
	This type of defect appears as a tiny surface damage on the animal skin, and is often neglected via human inspection.
	A sample defective image is shown in Figure~\ref{fig:sample_capture}.
	An instance segmentation deep learning model, namely, Mask Region-based Convolutional Neural Network (Mask R-CNN), is utilized to develop a robust architecture to evaluate the test dataset.
	Then, the details of the defective regions (i.e., a set of $XY$ coordinates) is transferred to a robotic arm to automatically mark the boundary of the defect area.

	In short, the contributions of this work are summarized as follows:
	
	\begin{enumerate}
		
		\item Proposal of an end-to-end defect detection and segmentation system using a deep convolutional neural network (CNN).
		
		\item Usage of robotic arm for automatic dataset collection and defect marking on the leather.
		
		\item Acquisition of a set of optimal $XY$ coordinates of each irregular shape of defect using mathematical derivation of the geometry.
		
		\item Thorough experiments on approximately \lst{80} training images and \lst{500} testing images and several performance metrics are presented to demonstrate the effectiveness of the proposed framework.
		
	\end{enumerate}
	
	The rest of the paper is organized as follows: 
	Section~\ref{sec:proposed} elaborates the details of our experimental setup and the proposed algorithm is thoroughly described.
	Section~\ref{sec:metric} explains the measurement metrics and the experiment settings.
	Section~\ref{sec:result}  reports and discusses the experimental results. 
	Lastly, Section~\ref{sec:conclusion} concludes the paper.
	
	\section{Proposed Method}
	\label{sec:proposed}
\begin{figure}[t!]
		\centering
		\includegraphics[width=0.8\linewidth]{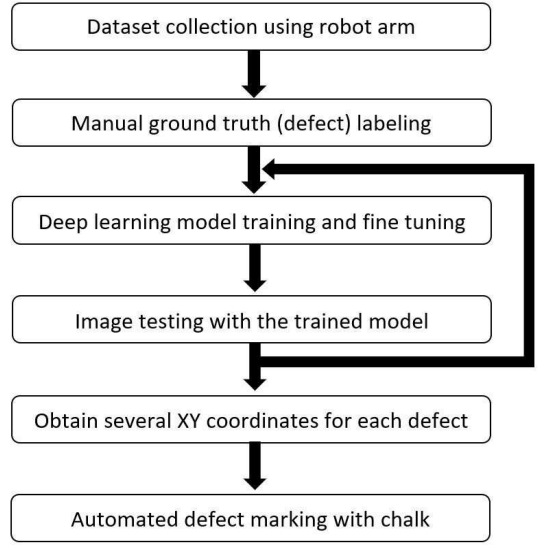}
		\caption{Flowchart of the proposed leather defect detection and segmentation framework}
		\label{fig:flow}
	\end{figure}

	The proposed automated visual defect inspection system comprised of six stages: 
	1) Dataset collection using robotic arm to capture top-down leather images;
	2) Manual ground truth annotation for each defect on all the images;
	3) Deep learning architecture training and parameters fine tuning on the trained model;
	4) Images testing with the trained model;
	5) Acquisition of a set of $XY$ coordinates for each defect;
	6) Defect highlighting with chalk using the robotic arm.
	The architecture overview of the system is illustrated in Figure~\ref{fig:flow}. 
	Following subsections explicitly describe each of the six steps involved.

	\subsection{Dataset collection using robotic arm}
\begin{figure}[t!]
		\centering
		\includegraphics[angle=270,origin=c,width=1\linewidth]{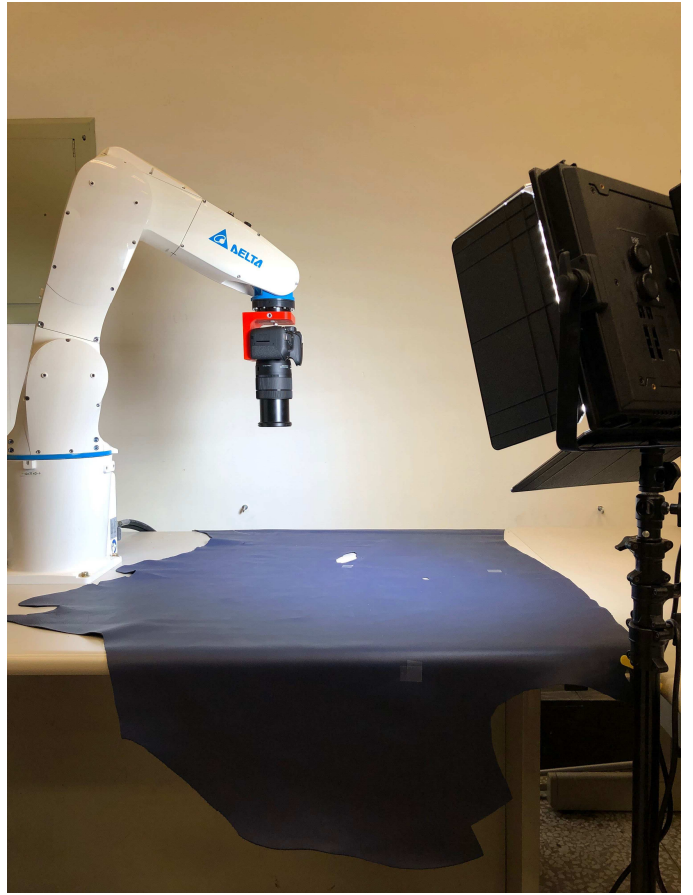}
		\caption{Hardware prototype of robot arm and lighting to capture the local image of leather}
		\label{fig:setup}
	\end{figure}

	The apparatus involved in the dataset elicitation consist of a six-axis desktop robotic arm, high-resolution camera, non flickering LED light source and 3D printed plastic components.
	The leather is placed on a table and there is a 2D camera mounted on a robot arm to capture the details of the leather from top-down viewpoint as illustrated in Figure~\ref{fig:setup}.
	The experiments are carried out on a six-axis articulated robots DRV70L from Delta.
	The placement of the Robot has been optimized, to reach the maximum range of the leather detection. 
	This optimization has been achieved by the commercial software package of Tecnomatix Process Simulation.
	The robot payload is 5kg and the weight of the camera is about 1kg. 
	All the movements of the robot have been programmed by the robot language of Delta Robot Automation Studio (DRAS), such that it will move to a few specific pre-configured positions to automatically capture multiple image patches.
	The control code can be transferred into the robot control gear by direct Ethernet link, and be independent from DRAS during the operation.
	In order to improve the image capture stability during the robot movement, the holding tool for the camera has been optimized.
	The images are captured using Canon EOS 77D and the detailed settings are described in Table~\ref{table:canon}.
	The fluorescent lights in the laboratory are operating on alternating current (50 Hz) electric systems produce light flickering, which yield to undesirable shadow, reflectance and variable illumination.
	Thus, a professional lighting source (i.e., DOF D1296 Ultra High Power LED light  of 12400 lux) is used to provide consistent and continuous source of illumination.
	Particularly, the light source is placed and fixed at 45 degrees from the leather and aiming downward.

	\setlength{\tabcolsep}{5pt}
	\begin{table}[tb]
		\begin{center}
			\caption{Camera specifications and configurations}
			\label{table:canon}
			\begin{tabular}{lc}
				\noalign{\smallskip}
				\hline
				\noalign{\smallskip}
				Feature
				& Description \\
				\hline
				\noalign{\smallskip}
				Megapixel 
				& 24.2 \\
				
				\noalign{\smallskip}
				Resolution (pixels)
				& 2400 $\times$ 1600 \\
				
				\noalign{\smallskip}
				Color representation
				& sRGB\\
				
				\noalign{\smallskip}
				Frame rate (fps)
				& 60\\
				
				\noalign{\smallskip}
				\lst{Shutter Speed (s)}
				& 1/60\\
				
				\noalign{\smallskip}
				Exposure time ($s$)
				& 1/200\\
				
				\noalign{\smallskip}
				ISO speed
				& ISO - 1600\\
				
				\noalign{\smallskip}
				Focal length (mm)
				& 135\\
				
				\noalign{\smallskip}
				Flash mode
				& No flash\\
				
				\hline
				
			\end{tabular}
		\end{center}
	\end{table}

	\subsection{Manual ground truth defect annotation}
	\label{subsec:manual}
	Each of the image collected is partitioned into a 400$\times$400 pixels.
	For instance, a 2400$\times$1600 pixel (equivalent to 90$\times$60$mm^2$) image will be divided into 24 pieces small image patches. 
	This is to allow the architecture designed in the later stage can better extract and learn the local features of leather effectively. 
	In addition, the partitioning step avoids the GPU computer from overloading by processing a high resolution image in parallel.
	We did not resize/ downsample the images, as the number of pixels in the image will be reduced and degrades the image quality.

\begin{figure*}[thpb]
		\centering
		\includegraphics[width=1\linewidth]{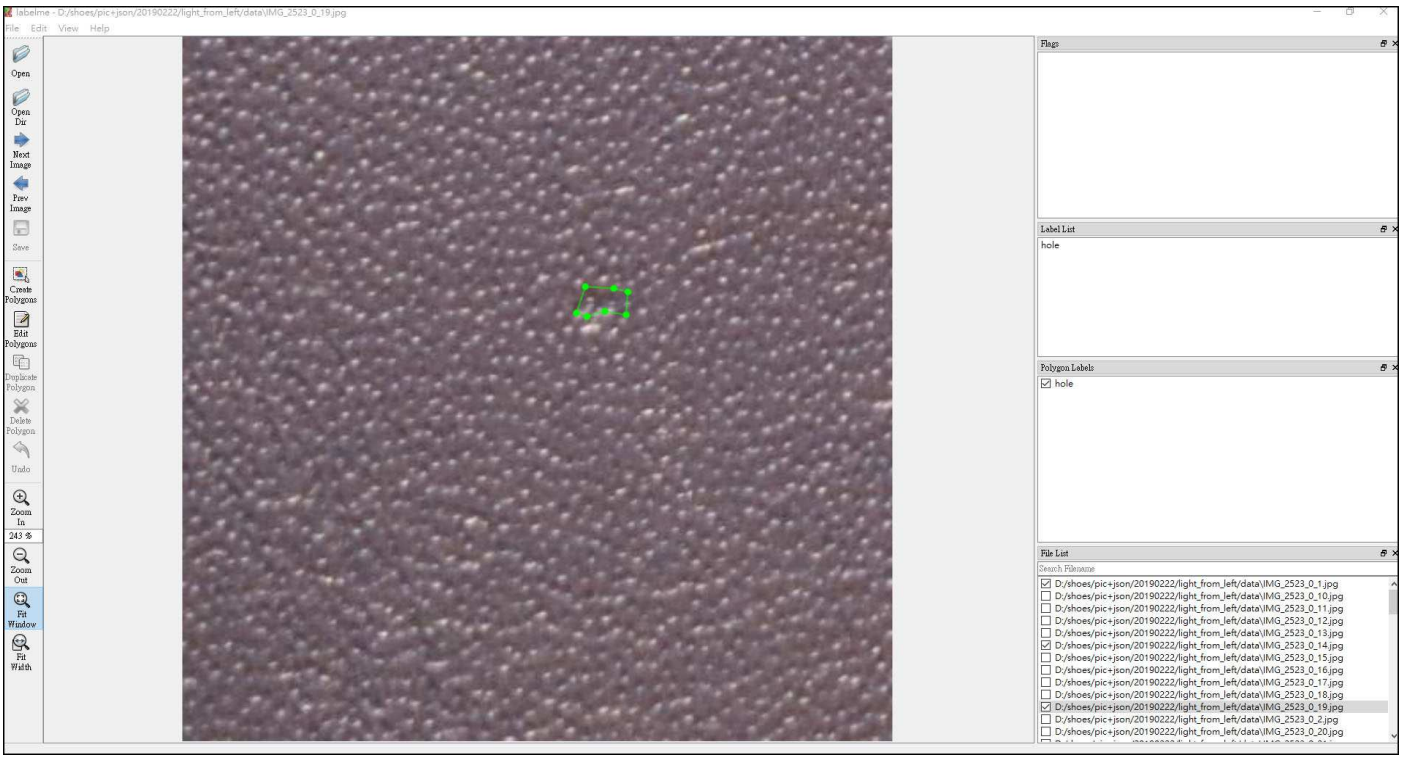}
		\caption{Manual ground truth labeling using an annotation tool}
		\label{fig:gt_label}
	\end{figure*}

\begin{figure*}[thpb]
		\centering
		\includegraphics[width=0.7\linewidth]{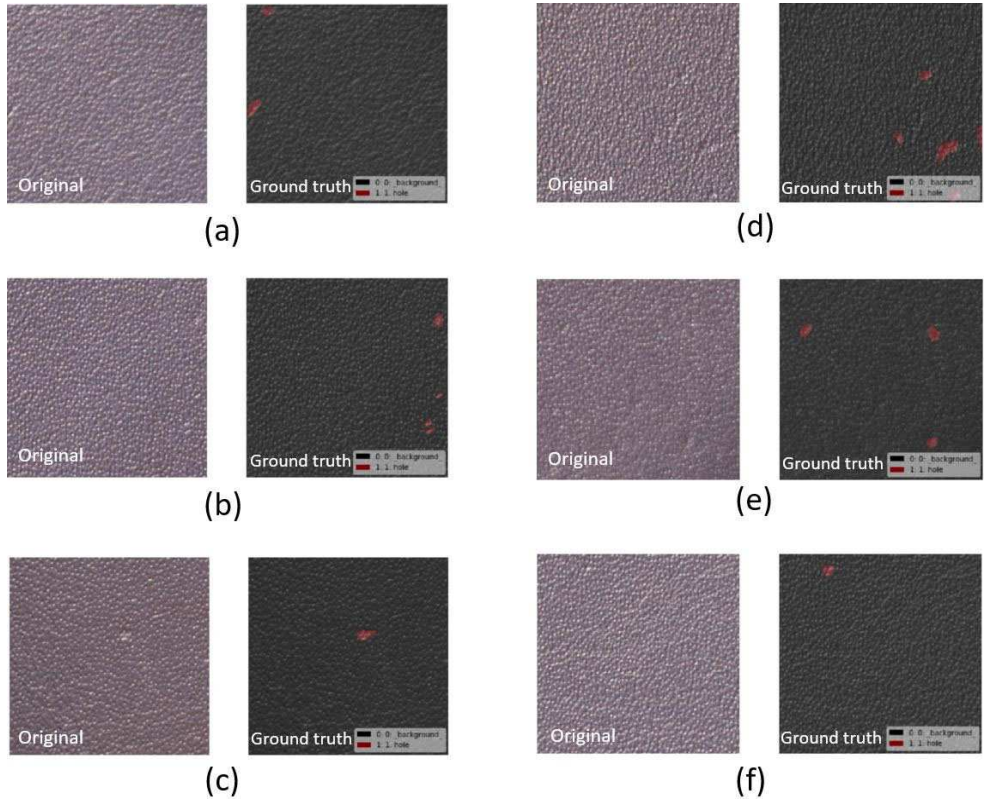}
		\caption{Sample image of six leather sheets, where the left is the original image captured by camera and the right is the visualization of groundtruth and background annotation.}
		\label{fig:gt}
	\end{figure*}

	The boundary of the defect region is annotated using an open source annotation tool~\cite{matterport_labelme_2016}, as depicted in Figure~\ref{fig:gt_label}.
	This software enables us to define the irregular shapes of the target object and assign a specific label name.
	The sample annotated binary masked image is illustrated in Figure~\ref{fig:gt} with reasonably precise pixel-wise closed boundary.
	Particularly, the left and the right images shown in Figure~\ref{fig:gt} are before and after performing the ground truth labeling, respectively.
	The red highlighted areas indicate the tick bite defect, whereas the black regions denote the background.
	To ensure the reliability of the ground truth labeled, five undergraduate students are involved in this step to assist in verifying the correctness of the labels.

	\subsection{Deep learning model training and fine tuning}
	
	There is a total of \lst{584} images that had been performing the annotation process in the previous step (i.e., Section~\ref{subsec:manual}).  
	The dataset is thereafter apportioned into training and testing sets, with an \lst{15 - 85 split}.  
	Concisely, there are \lst{84} images categorized as the training dataset and 500 images as the testing dataset.
	The architecture exploited to learn the features of the defective and non-defective regions is the Mask R-CNN  (Regional Convolutional Neural Network) .
	It is a popular image segmentation model that built  Feature Pyramid Network (FPN)~\cite{lin2017feature} with a ResNet-101~\cite{he2016deep} backbone.
	
\begin{figure}[t!]
		\centering
		\includegraphics[width=1\linewidth]{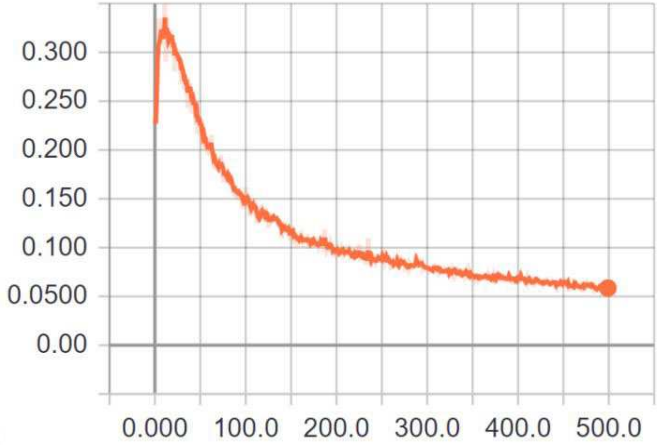}
		\caption{Training loss for the defective region using Mask R-CNN (x-axis denotes the epoch number and y-axis is the training loss)}
		\label{fig:loss}
	\end{figure}

	Mask R-CNN model has been pre-trained extensively on a Microsoft Common Objects in Context dataset (MS COCO)~\cite{lin2014microsoft}, which incorporate over 1.5 million segmented images from a total of 80 categories.  
	On top of performing the transfer learning from the pre-trained model to detect and segment the defects of the leather, the parameters (i.e., weights and biases) are iteratively adjusted through learning the features of the leather input images. 
	As such, the pre-trained model with rich feature extraction can be tuned to improve the recognition performance in recognizing the leather defect.  
	As a result, the model is capable to learn rich feature representations quickly, which is certainly better than training a naive network with randomly initialized weights from scratch.
	The model training process is conducted on two GPUs (Graphics Processing Unit) for \lst{400} epochs and it is completed in 7 hours.
	The model is trained with Adam optimizer with a learning rate of 0.001.
	Figure~\ref{fig:loss} shows the training accuracy for each epoch.
	It can be seen that the training loss gradually decreases at about 5\% after epoch = 400.
	
	The details of the architecture and setting values are listed in Table~\ref{table:configuration}.
	In the code implementation, there is a Region Proposal Network (RPN), which is a lightweight neural network that scans the image in a sliding-window fashion and searches for the areas that contain targeted objects.
	We did a slight modification on the RPN anchor scales, where a range of more than 33\% of positive ROIs has been chosen as the sampling among the entire image. 
	This is because the targeted object (i.e., defective area) in our experiment is very small.
	

	\setlength{\tabcolsep}{5pt}
	\begin{table}[tb]
		\begin{center}
			\caption{Configuration type and Parameters of Mask R-CNN model}
			\label{table:configuration}
			\begin{tabular}{lc}
				\noalign{\smallskip}
				\hline
				\noalign{\smallskip} Configuration type & Parameter \\
				\hline
				
				\noalign{\smallskip} Backbone strides    & $[4, 8, 16, 32, 64]$ \\
				\noalign{\smallskip} Batch size   & 2 \\
				\noalign{\smallskip} bbox standard deviation     &  $[0.1 0.1 0.2 0.2]$\\
				\noalign{\smallskip} Detection Max Instances      &    100\\
				\noalign{\smallskip} Detection Min Confidence     &    0.7\\
				\noalign{\smallskip} Detection Nms Threshold      &    0.3\\
				\noalign{\smallskip} Fpn Classif Fc Layers Size   &    1024\\
				\noalign{\smallskip} Gpu Count                  &      2\\
				\noalign{\smallskip} Gradient Clip Norm          &     5.0\\
				\noalign{\smallskip} Images Per Gpu             &      1\\
				\noalign{\smallskip} Image Max Dim              &      512\\
				\noalign{\smallskip} Image Meta Size            &      14\\
				\noalign{\smallskip} Image Min Dim              &      512\\
				\noalign{\smallskip} Image Min Scale           &       0\\
				\noalign{\smallskip} Image Resize Mode          &      square\\
				\noalign{\smallskip} Image Shape                 &     $[512~512~3]$\\
				\noalign{\smallskip} Learning Momentum          &      0.9\\
				\noalign{\smallskip} Learning Rate             &       0.001\\
				\noalign{\smallskip} Mask Pool Size             &      14\\
				\noalign{\smallskip} Mask Shape                 &      [28, 28]\\
				\noalign{\smallskip} Max Gt Instances           &      100\\
				\noalign{\smallskip} Mean Pixel                 &      [123.7 116.8 103.9]\\
				\noalign{\smallskip} Mini Mask Shape             &     (56, 56)\\
				\noalign{\smallskip} Name                       &      shapes\\
				\noalign{\smallskip} Num Classes                 &     2\\
				\noalign{\smallskip} Pool Size                  &      7\\
				\noalign{\smallskip} Post Nms RoIs Inference    &      1000\\
				\noalign{\smallskip} Post Nms RoIs Training      &     2000\\
				\noalign{\smallskip} RoI Positive Ratio         &      0.33\\
				\noalign{\smallskip} RPN Anchor Ratios         &       $[0.5, 1, 2]$\\
				\noalign{\smallskip} RPN Anchor Scales         &       $32, 32, 48, 48, 96)$\\
				\noalign{\smallskip} RPN Anchor Stride         &       1\\
				\noalign{\smallskip} RPN bbox Std Dev           &      $[0.1 0.1 0.2 0.2]$\\
				\noalign{\smallskip} RPN NMS Threshold          &      0.7\\
				\noalign{\smallskip} RPN Train Anchors Per Image  &    256\\
				\noalign{\smallskip} Steps Per Epoch              &    100\\
				\noalign{\smallskip} Top Down Pyramid Size       &     256\\
				\noalign{\smallskip} Train BN                  &       False\\
				\noalign{\smallskip} Train RoIs Per Image      &       33\\
				\noalign{\smallskip} Use Mini Mask             &       True\\
				\noalign{\smallskip} Use RPN RoIs              &       True\\
				\noalign{\smallskip} Validation Steps         &        20\\
				\noalign{\smallskip} Weight Decay              &       0.0001\\
				
				\hline
			\end{tabular}
		\end{center}
	\end{table}

	\subsection{Images testing with the trained model}
	
\begin{figure}[t!]
		\centering
		\includegraphics[width=1\linewidth]{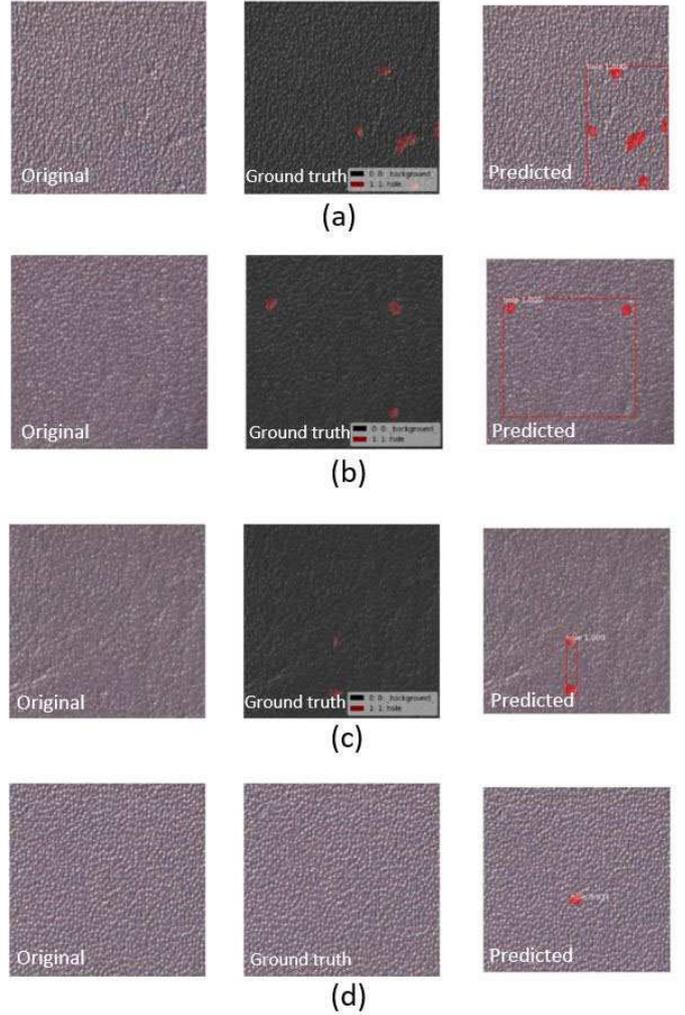}

		\caption{Example of visualization to compare the original image (left), manual annotated image (middle) and predicted results using the trained model (right) }
		\label{fig:predicted}
\end{figure}

	The test images are fed into the Mask R-CNN to extract the region proposals, that are high potential areas to be the target object.
	There are two sets of fully-connected layers at the end of the Mask R-CNN to perform both the classification (i.e., class label predictions) and instance segmentation (i.e., pixel-wise box locations for each predicted object).
	The sample output of a testing image is shown in Figure~\ref{fig:predicted}.
	The left column of Figure~\ref{fig:predicted} shows the original leather, middle column is the manually annotated ground truth with human effort, and the right column is the test result generated using the trained model.
	It can be seen that the right column comprised of the  boundary boxes of uneven shapes with its predicted label.
	Specifically, the trained model in Figure~\ref{fig:predicted} predicts: 
	(a) all the five defects correctly; 
	(b) two defects correctly and misses a defect;
	(c) two defects correctly, and;
	(d) a false alarm;

\subsection{\lst{Acquisition several $XY$ coordinates for each defect}}

For each defect detected by the Mask R-CNN, instead of considering \textit{all} points on the defect boundary, we opt to select \textit{a few} $XY$ coordinates of the points, which will be served as the input into the robot arm for the defect marking process. 
This points selective process is called the \textit{boundary optimization}. 
The intention to reduce the number of points is that the robot arm moves with very tiny steps (down to 0.03mm), and hence not every single pixel on the defect boundary mask are required to be marked as the defect. 
Practically, the number of the $XY$ coordinates of a defect boundary might be in the range of \lst{200 to 2000}.
Thus, it leads to an overloading phenomena on the robotic arm if all the points of the boundary are considered. 

\begin{figure}[t!]
	\centering
	\includegraphics[width=1\linewidth]{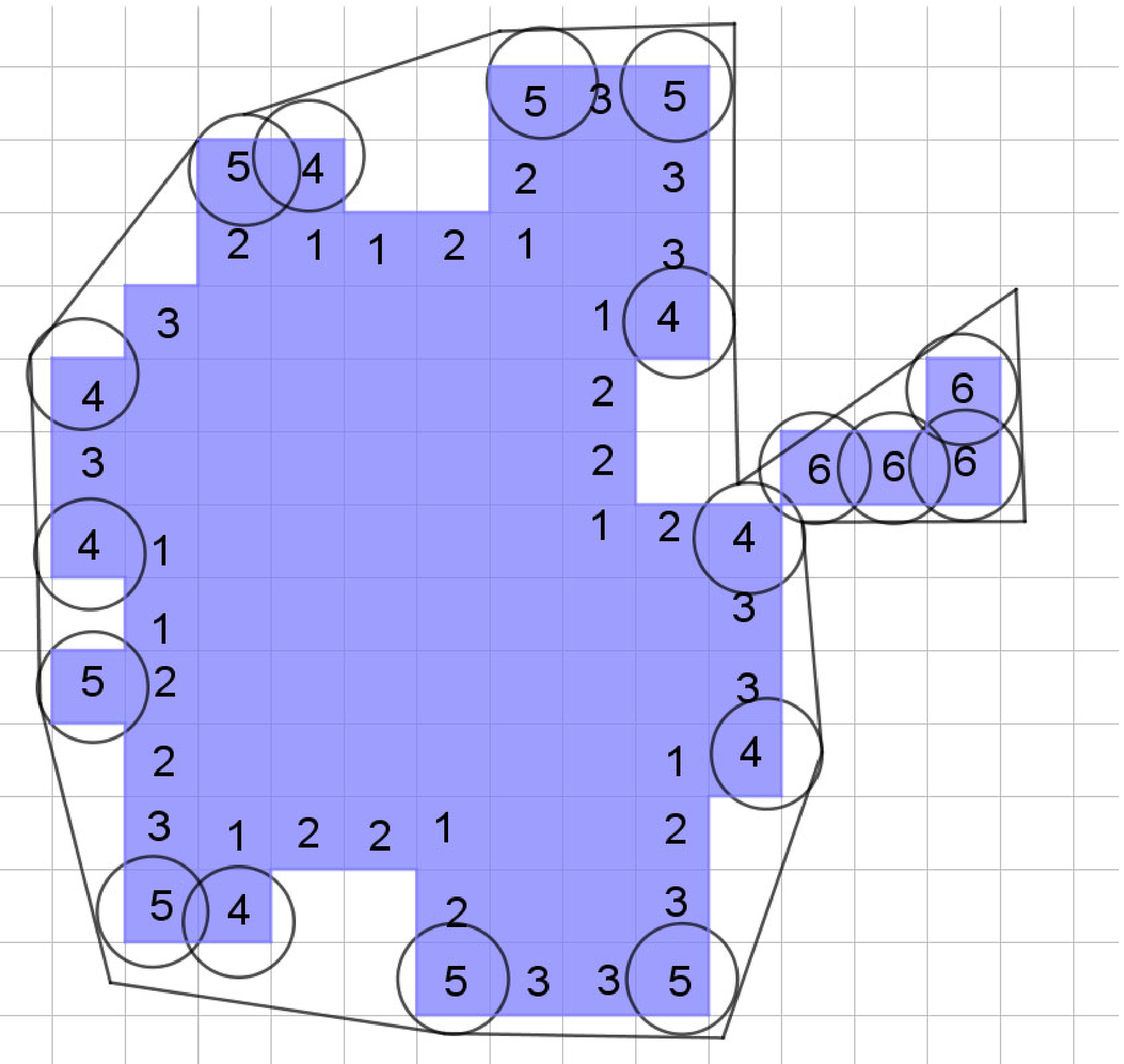}
	\caption{Points with vacuum $\geq 4$ preserve the geometric properties: $6$ represents the linking points, $4,5$ are the corners to maintain the convexity of the defect.}
	\label{fig:lbp0}
\end{figure}

\begin{figure*}[t!]
	\centering
	\includegraphics[width=1\linewidth]{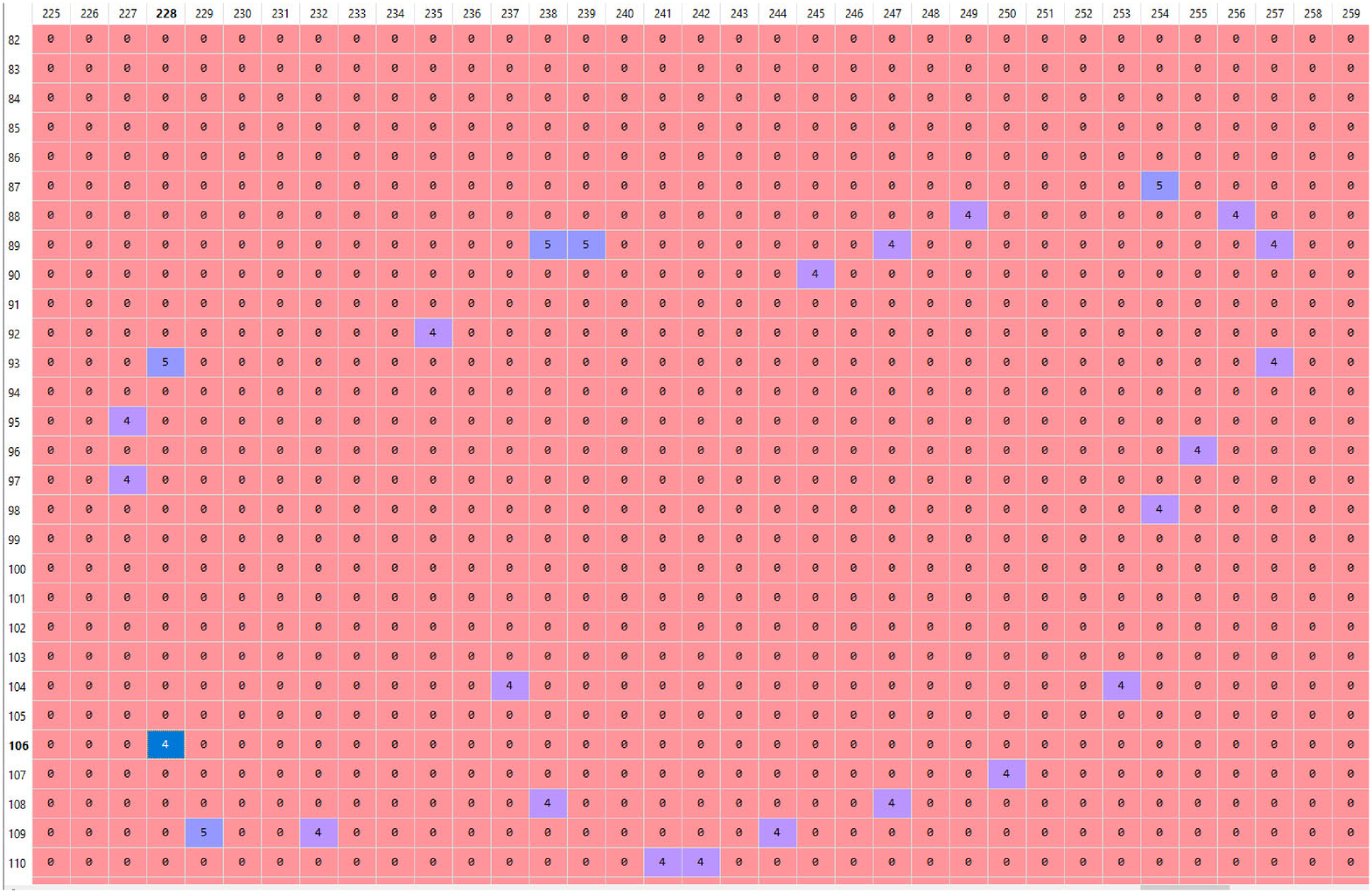}
	\caption{Final selected points which consists of values 4 or 5, highlighted in blue.}
	\label{fig:lbp}
\end{figure*}

In order to achieve the boundary optimization, the criterion of choosing ``salient'' $XY$ coordinates of points should satisfy these three rules: 
(1) The selected points lies on the outer boundary; 
(2) The number of the points should be as little as possible to sufficiently represent the original boundary; 
(3) The polygon formed by connecting the selected points preserves the geometrically significant properties of the boundary, including the corners of the boundary and the convexity. 
Note that, for each defect, the number of selected points is different as it depends on its shape and size.
As a consequence, a few steps have been proposed to obtain the significant points, as elaborated as follows:

\begin{enumerate}
	\item \textbf{Determination of the region for each defect}\\
	From the predicted masks shown in Figure~\ref{fig:predicted}, the red highlighted regions are the predicted defect mask which comprised of ``TRUE" value all the pixels within the area, while the rest of the pixels consist of ``FALSE" value.
	All the values are then binarized to become either 1 or 0 by function $g$  as defined in Equation \ref{eq:1_0}:
	\begin{equation}
	\label{eq:1_0}
	g(x) = \begin{dcases*}
	1,  & x = TRUE\\
	0, & x = FALSE
	\end{dcases*}
	\end{equation}
	
	The example of the binarized image is shown in Figure~\ref{fig:shape of predicted mask}.\\
	
	\begin{figure*}[thpb]
		\centering
		\includegraphics[width=1\linewidth]{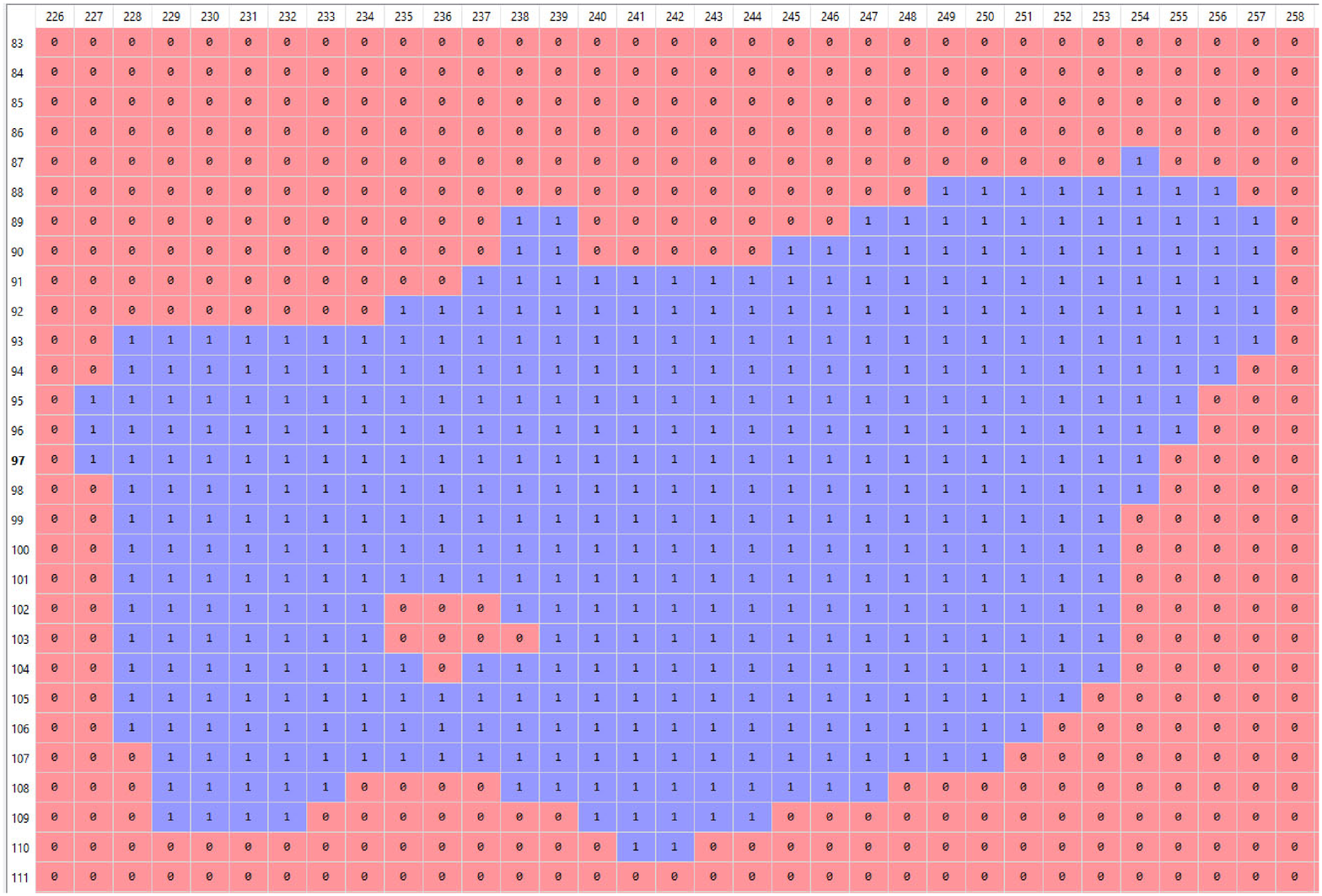}
		\caption{Example of a binarized image, where the defective region is indicated by value 1, while the rest of the pixels indicate there is no defect}
		\label{fig:shape of predicted mask}
	\end{figure*}

	\item \textbf{Acquisition of the defect boundary}\\   
	A modified local binary coding is applied on the binarized image to derive a new image representation.
	The original Local Binary Pattern~\cite{ojala1996comparative} feature descriptor has been widely applied in the computer vision due to its advantage of: (a) discrimination ability; (b) compact texture representation; (c) low computational complexity, and; (d) invariant to any monotonic gray-level changes.
	In brief, LBP operator compares the intensity value of the center pixel to its circular neighboring pixels using a thresholding technique.
	Specifically, given a pixel c at position ($x_c$ , $y_c$), the binary code is computed by comparing the value of pixel c with its neighboring pixels:  
	
	\begin{equation}
	\label{eq:lbp}
	LBS_{P,R} = \sum\limits_{p=0}^{P-1} s(g_c - g_p) , s(x) = \begin{dcases*}
	1,  & x $>$ 0\\
	0, & x $\leq$ 0
	\end{dcases*}
	\end{equation}
	
	\noindent 
	where $P$ is the number of neighbouring points around the center pixel, $(P,R)$ representing a neighbourhood of $P$ points equally spaced on a circle of radius $R$, $g_c$ is the gray value of the center pixel and $g_p$ are the $P$ gray values of the sampled points in the neighbourhood.
	
	For instance, if all the 8 neighbouring pixels are zeros in a $3 \times 3$ pixel grid and the middle pixel is 1, the value of the center in the grid will be replaced by the value of 8, as shown in the left image in Figure~\ref{fig:vacuum}.
	In contrast, if the pixel point is an isolated point, its vacuum value would be $0$. 
	For clarity, the notations used to describe this new derived value is defined as \textit{vacuum}, $V$. 
	$V$ is always an integer and $0\leq V\leq 8$.  
	Theoretically, unique and important boundary points are indicated by a large $V$ value, and vice versa.
	The image that has been carried out the modified LBP process is illustrated in Figure~\ref{fig:shape}.\\
	
	\begin{figure}[t!]
		\centering
		\includegraphics[width=1\linewidth]{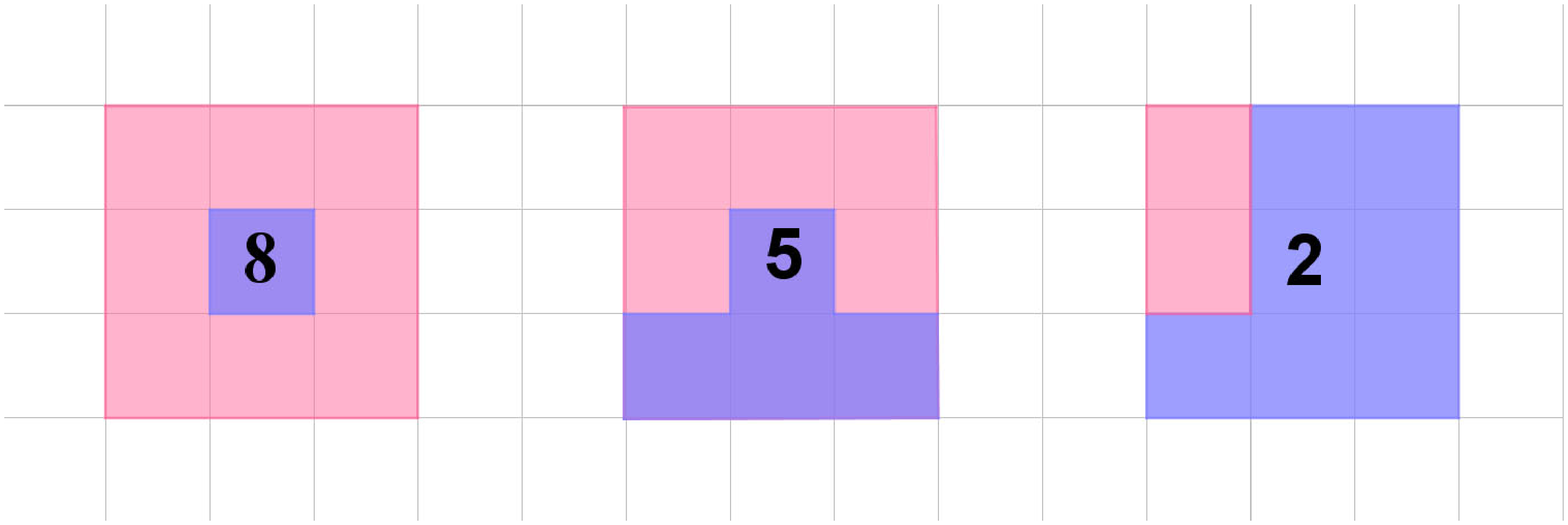}
		\caption{The vacuum, $V$ = 8, 5, 2 (from left to right). The red pixels are the pixel with 0 while the blue pixels contain 1.}
		\label{fig:vacuum}
	\end{figure}

	\begin{figure*}[t!]
		\centering
		\includegraphics[width=1\linewidth]{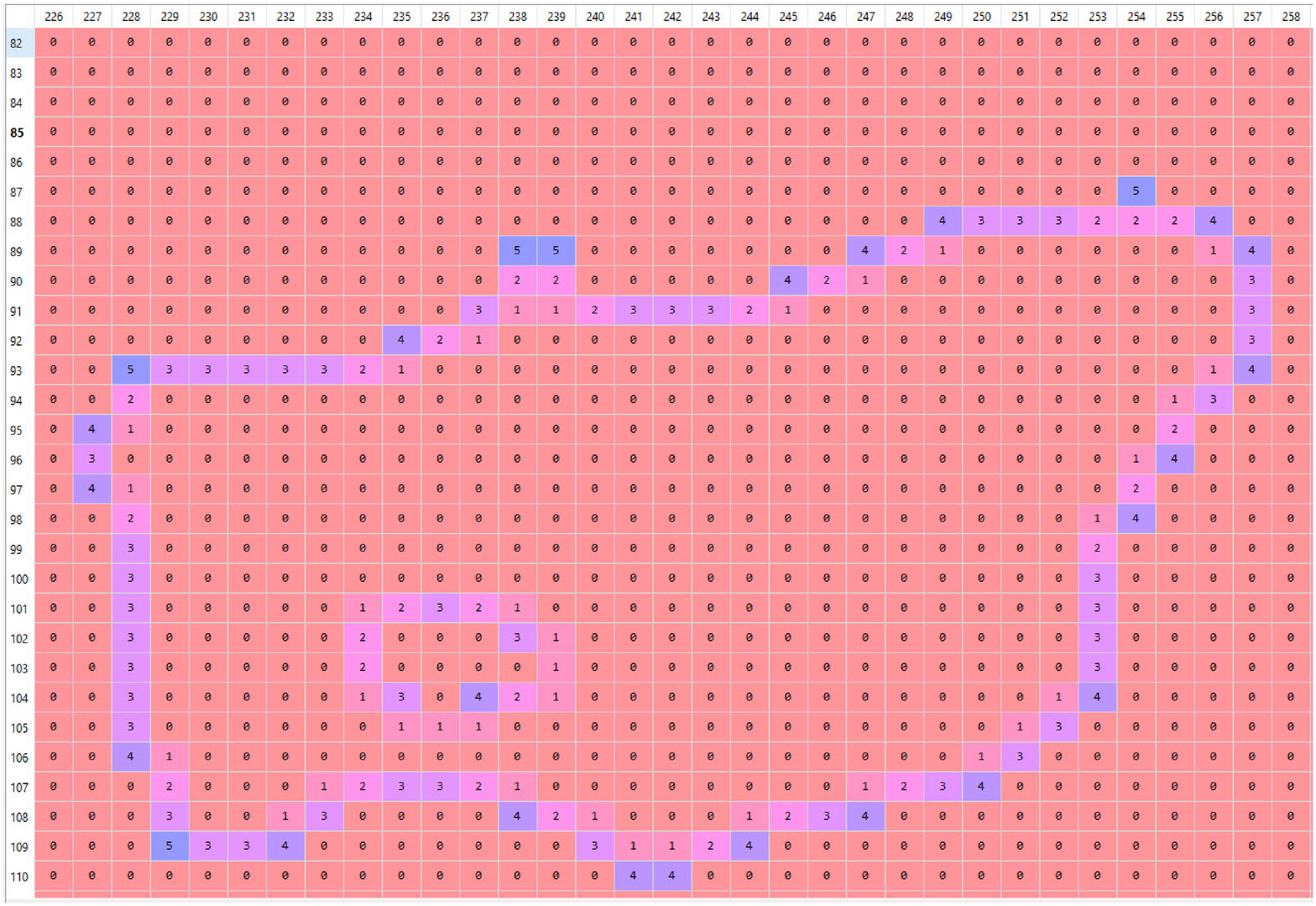}
		\caption{Example of mask shape: after obtaining the vacuum $V$ on the boundary}
		\label{fig:shape}
	\end{figure*}

	\item \textbf{Extraction of a set of geometrically significant points}

	According to the point-set topology, a boundary point (i.e., $p$) from a set (i.e., $S$) on a plane is defined such that for any solid disk (i.e., $D$) centered at $p$ with any radius, $D$ must contain some points in and out of the set $S$ at the same time.
	In the discrete case, a boundary point consists of the vacuum value of $4$, which is the midpoint between $0$ and $8$. 
	To select the important points,  only the points with $V\geq 4$ is considered. 
	This is because these points are most likely the corner, convexity, or the linking points of the defect boundary.
	Figure~\ref{fig:lbp0} clearly illustrates the selected boundary points with their geometric quantities and vacuums.
	However, since most of the shape of the defective masks are closed to an elliptic shape, the boundaries points with anomaly shapes will be neglected.  
	In the experiments conducted in this paper, only the points where $V = \{4,5\}$ are selected, as $V = \{6, 7\}$ are most likely the defective masks with spikes, which may be the noise or irrelevant to the defects.
	The final selected points are shown in Figure~\ref{fig:lbp}, which consists of values 4 or 5, and are highlighted in blue.
	
\end{enumerate}

\begin{figure}[t!]
	\centering
	\includegraphics[width=1\linewidth]{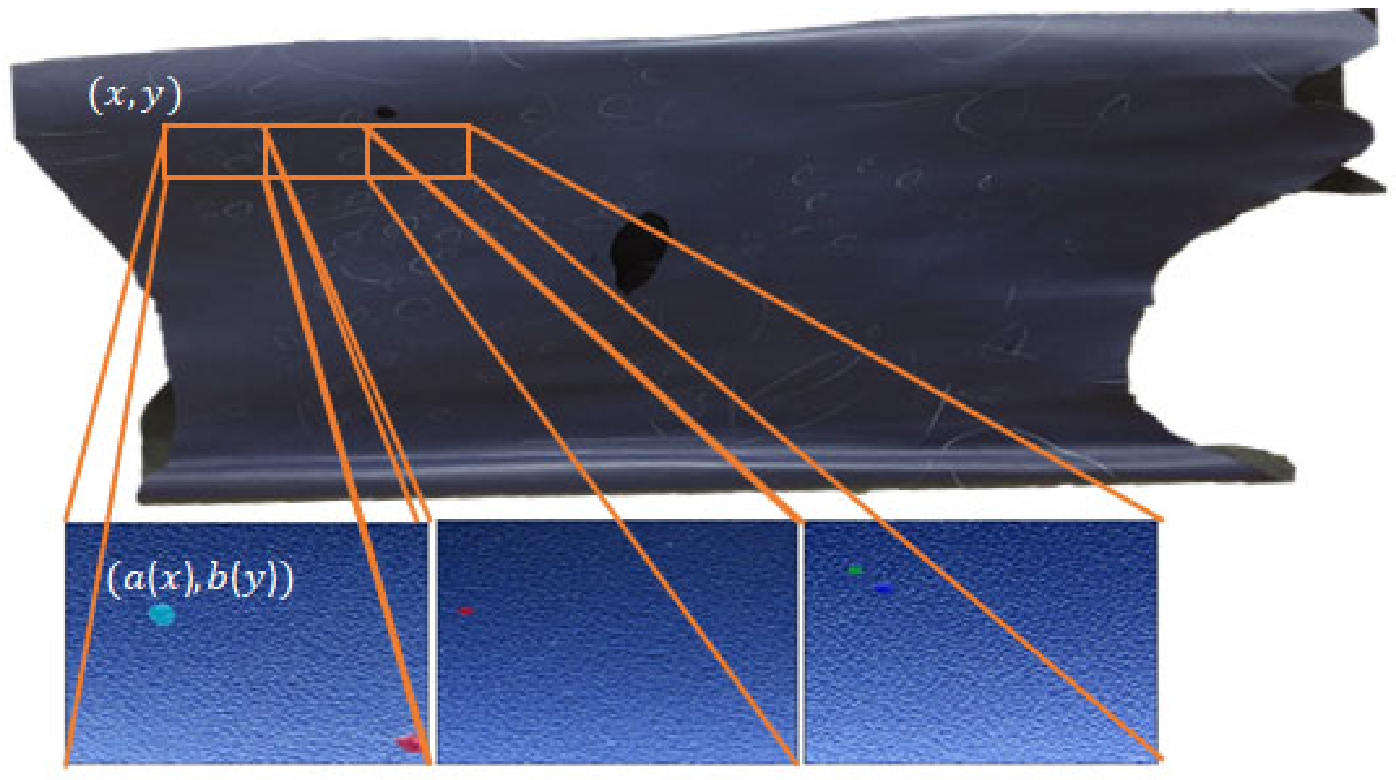}
	\caption{Example of mapping for the images captured to the actual leather}
	\label{fig:projection}
\end{figure}

To form a continuous bounding mask for each defect, all the selected points are connected in counterclockwise direction using the Graham Scan algorithm~\cite{graham}.
Generally, this algorithm creates a convex polygon that contains all the selected points, at the meantime ignores the inner boundary points.
The selected points are then transformed into corresponding coordinate in real space by using a simple pixel transformation method based on the scale of the image resolution captured and the physical leather size. 
For example, given that the reference coordinate of an actual image is $(x_0, y_0)$. 
For each defective $X,Y$ coordinates derived from the previous steps $(a_i, b_i)$, the actual physical defect coordinate (i.e., in millimeter) is computed as follows:
\begin{equation}
(x_i, y_i) = (x_0 + \omega_1 x_i ,  y_i = y_0 + \omega_2 y_i) ,
\end{equation}
where $\omega_1$ and $\omega_2$ are the projection ratio for the width and height of leather, respectively.
For better visualization, Figure~\ref{fig:projection} shows the example of the mapping for the images captured to the actual leather.

\subsection{Automated defect marking with chalk}

\begin{figure}[t!]
	\centering
	\includegraphics[angle=0,origin=c,width=0.7\linewidth]{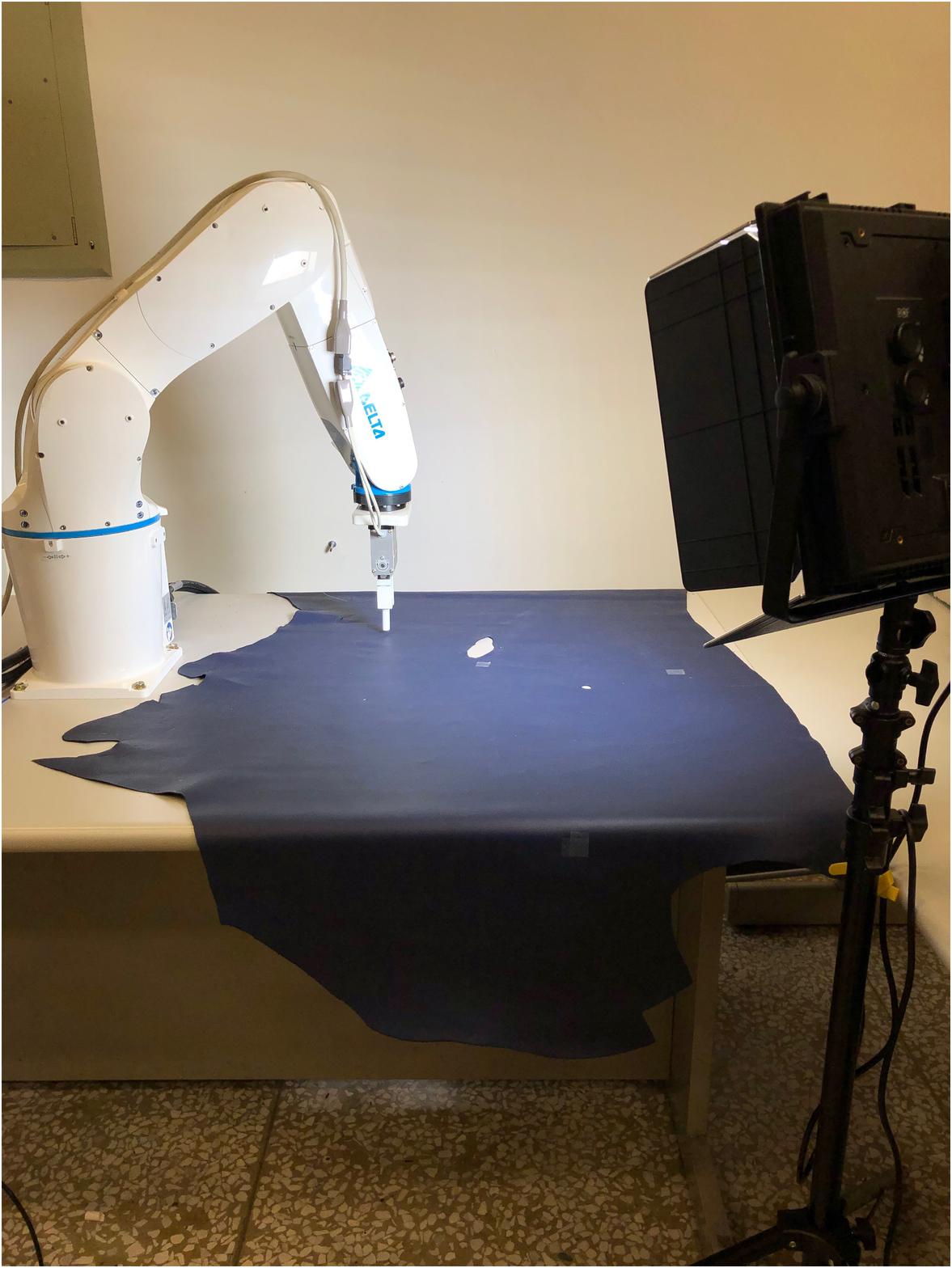}
	\caption{Defect boundary marking using robotic arm}
	\label{fig:defect_marking}
\end{figure}

In this step, the robot arm is moved and the defect area is marked according to the sets of the coordinate produced from the previous step. 
Specifically, the camera that mounted on the robot arm is replaced by an erasable liquid chalk.
The example of the defect marking process is illustrated in Figure~\ref{fig:defect_marking}.

\section{Metrics and Experiment Settings}
\label{sec:metric}

There are six evaluation metric to validate the performance of the proposed framework, namely, specificity, sensitivity, precision, F1-score, error rate and accuracy:

\begin{equation}
\text{Sensitivity} := \frac{\text{TP}}{\text{TP + FN}}, 
\end{equation}

\begin{equation}
\text{Specificity} := \frac{\text{TN}}{\text{TN + FP}}, 
\end{equation}

\begin{equation}
\text{Precision} := \frac{\text{TP}}{\text{TP + FP}},
\end{equation}

\begin{equation}
\text{F1-score} := 2 \times \frac{\text{Precision} \times \text{Specificity}}{\text{Precision + Specificity}}, 
\end{equation}

\begin{equation}
\text{Error rate} := \frac{\text{FP + FN}}{\text{TP+TN+FP+FN}},
\end{equation}

\begin{equation}
\text{Accuracy} := \frac{\text{TP + TN}}{\text{TP + FP + TN + FN}}
\end{equation}

\noindent where TP, TN, FN and FP are the true positive, true negative, false negative and false positive, respectively.
The definition of these four terms are explained as follows:
\begin{itemize}
	\item TP: The model correctly segments the tick bite defect.
	\item TN: The outcome where the model correctly predict there is no defective area.
	\item FN: The event where the model does not segment the defect correctly, while in fact there is a defect. 
	\item FP: The test result indicates there is a defect exists, but there is none. This phenomena is commonly called a ``false alarm".
\end{itemize}
Specificity and sensitivity refer to the effectiveness of the algorithm to identify the defective and non-defective areas, respectively.
Precision is the measure of how much information in the system is returned correctly.
F1-score is the weighted average of the specificity and precision.
The error rate indicates the percentage of the number of incorrect segmented instances, while accuracy is the measure of the degree of closeness of the predicted output to the ground truths.

\section{Experimental Results and Discussion}
\label{sec:result}

All the experiments were carried out on Python 3.6 in Intel Core i7-8700K 3.70 GHz processor, RAM 32.0 GB, GPU NVIDIA GeForce GTX 1080 Ti.

The performance of the defect instances segmentation is shown in Table~\ref{table:result}. 
Both the train and test datasets utilize the same trained Mask R-CNN architecture to predict the output.
The number of sample images in the train and test datasets are 84 and \lst{500}, respectively.
It is observed that the accuracy for train dataset is higher than the test one.
Particularly, the model exhibits accuracies of 91.5\% for the training data and 70.35\% for the testing data.
Theoretically, the test accuracy is less than of the train accuracy.
This is due to the test data is unseen by the trained model, and train data is exactly the data used to train the model.
On the other hand, the specificity and the F1-score for the train dataset in Table~\ref{table:result} are zeros.
This is because there will be no TN case in evaluating the training data, since the training images are restricted to that comprised of defective area.
In other words, non-defective images are not applicable to be the training data.
In addition, the specificity in the test dataset in Table~\ref{table:result} is 75.81\%, this implies majority of the testing images are non-defective images, and the model is capable to predict them correctly. 

To further analyze the performance, confusion matrices are provided in Table~\ref{table:cf} and \lst{Table~\ref{table:cf2}}, for the train and test datasets, respectively.
In brief, confusion matrix is a typical measurement to illustrate the classification rate for each defective or non-defective cases.
It can be seen in Table~\ref{table:cf},  there are 97 of defective regions correctly spotted.
Although the total number of images is 84, it should be reminded that one image may consist of more than one defect (as shown in Figure~\ref{fig:gt}).
As in Table~\ref{table:cf2}, there are 104 cases indicating the model is not able to detect the defective areas correctly.
However, due to the high TN value (i.e., 326), it leads to a reasonable overall testing accuracy.

\setlength{\tabcolsep}{5pt}

\begin{table}[tb]
	\begin{center}
		\caption{Leather segmentation performance on train and test datasets}
		\label{table:result}
		\begin{tabular}{lcc}
			
			\noalign{\smallskip}
			\cline{2-3} 
			\noalign{\smallskip}

			& Train dataset (\%)	 & Test dataset (\%)	\\
			
			\noalign{\smallskip}
			\hline
			\noalign{\smallskip}

			Sensitivity  
			& 97.00	 & 	53.57	\\
			
			Specificity 
			& 0	 &  75.81 \\

			Precision  
			& 94.17	 & 	41.90	\\
			
			F1-score 
			& 0 &  53.97 \\
			
			Error rate  
			& 8.50 & 29.65		\\
			
			Accuracy
			& 91.50	 & 70.35 \\
			
			\hline
		\end{tabular}
	\end{center}
\end{table}

\setlength{\tabcolsep}{5pt}

\begin{table}[tb]
	\begin{center}
		\caption{Confusion matrices of the proposed detection and segmentation system on the train dataset, which consists of 84 images}
		\label{table:cf}
		\begin{tabular}{lcccccc}
			
			\noalign{\smallskip}
			\cline{3-4} 
			\noalign{\smallskip}
			
			& & \multicolumn{2}{c}{Predicted}\\
			
			\noalign{\smallskip}
			\cline{3-4} 
			\noalign{\smallskip}
			
			& & Defective	 & Non-defective	\\
			
			\noalign{\smallskip}
			\hline
			\noalign{\smallskip}
			
			\multirow{2}{*}{Actual}
			
			& Defective  
			& 97	 & 6		\\
			
			& Non-defective 
			& 3	 & 0 \\
			
			\hline
		\end{tabular}
	\end{center}
\end{table}

\setlength{\tabcolsep}{5pt}

\begin{table}[tb]
	\begin{center}
		\caption{Confusion matrices of the proposed detection and segmentation system on the test dataset, which consists of 500 images}
		\label{table:cf2}
		\begin{tabular}{lcccccc}
			
			\noalign{\smallskip}
			\cline{3-4} 
			\noalign{\smallskip}
			
			& & \multicolumn{2}{c}{Predicted}\\
			
			\noalign{\smallskip}
			\cline{3-4} 
			\noalign{\smallskip}
			
			& & Defective	 & Non-defective	\\
			
			\noalign{\smallskip}
			\hline
			\noalign{\smallskip}
			
			\multirow{2}{*}{Actual}
			
			& Defective  
			& 75	 & 104		\\
			
			& Non-defective 
			& 65	 & 326 \\
			
			\hline
		\end{tabular}
	\end{center}
\end{table}

\section{Conclusion}
\label{sec:conclusion}
This paper presents an automatic approach to inspect the defect instances of a leather, yielding promising segmentation accuracies of \lst{91.50\%} and \lst{70.35\%} on the train and test datasets, respectively.
A specific defect type is focused in this study, namely the tick bite. 
Concretely, the proposed framework capture the image each local regions of the leather using a robotic arm.
Then, a ground truth labeling step is performed on the defective areas, which will be served as the input information to train the Mask R-CNN deep learning architecture.
Test images are fed into the trained model to acquire the $XY$ coordinates for each predicted defect area.
A modified LBP method is introduced to highly reduced the amount of the defective boundary points whilst maintaining the shape of the defect.
Finally, the robot arm is utilized to automatically sketch the defect boundaries with a erasable chalk, based on the coordinates derived.
Throughout the entire process, there is one step that requires human intervention: the ground truth annotation.
However, it should be noted that this step only performs once for the same type of leather.

As the future works, a mechanism can be designed to roll the leather automatically, instead of involving human assistance in such a case where the robotic arm cannot reach certain distance.
In addition, the trained Mask R-CNN architecture can be tested on the leather with different shape, color and texture to evaluate the robustness of the proposed approach.
Last but not least, rather than segmenting the tick bite defect, the same procedure can be evaluated for other defects, such as wrinkles, cuts, scabies, brand marks made from hot iron, etc.

\section*{Acknowledgments}
This work was funded by Ministry of Science and Technology (MOST) (Grant Number: MOST 107-2218-E-035-016-).

\end{document}